\documentclass[11pt]{article}

\usepackage[preprint]{acl}

\usepackage{times}
\usepackage{latexsym}
\usepackage{booktabs}
\usepackage{array}
\usepackage{tabularx}
\usepackage{siunitx}
\usepackage{multirow}
\usepackage{float}
\usepackage{kotex}
\usepackage{amsmath}
\usepackage{amssymb}
\usepackage{caption}
\usepackage{makecell}

\usepackage[T1]{fontenc}

\usepackage[utf8]{inputenc}

\usepackage{microtype}

\usepackage{inconsolata}

\usepackage{graphicx}

\title{
Multi-Perspective Triad Interaction Graph Neural Network \\
for Cognitive Distortion Detection
}

\author{
\textbf{Jun Seo Kim}\textsuperscript{1}
\qquad
\textbf{Hye Hyeon Kim}\textsuperscript{2}
\\
\textsuperscript{1}Department of Computer Engineering, Gachon University
\\
\textsuperscript{2}Department of Biomedical Systems Informatics, Yonsei University
\\
\texttt{kma80kjs@gachon.ac.kr}
\\
\texttt{hye\_hyeon@yonsei.ac.kr}
}

\begin{document}
\maketitle

\begin{abstract}
Cognitive distortion detection is a key task in computational mental health, yet existing approaches often overlook the psychological structure of distorted thoughts. We propose MTI-GNN (Multi-Perspective Triad Interaction Graph Neural Network), which models Beck's cognitive triad---negative views of the self, world, and future---as complementary perspectives for classification. An LLM decomposes each utterance into the three perspectives, from which perspective-specific similarity graphs are constructed and encoded by a Multi-Perspective GNN. A Triad Interaction module models cross-perspective dependencies through sequential source-conditioned updates and feature-wise gating, while Prototype-Guided Perspective Fusion performs label-conditioned aggregation. Label-expanded supervision incorporates all available distortion annotations during training. We evaluate MTI-GNN on 9,764 samples from four Korean, English, and Chinese datasets spanning ten distortion categories. MTI-GNN significantly outperforms all supervised variants and exceeds eight prompted generative models under zero-shot and few-shot settings. Leave-one-perspective-out ablations show that all three perspectives contribute significantly, while human expert evaluation provides preliminary evidence of their alignment with the intended cognitive dimensions.
\end{abstract}

\section{Introduction}

\begin{figure}[t]
\centering
\includegraphics[width=0.75\linewidth]{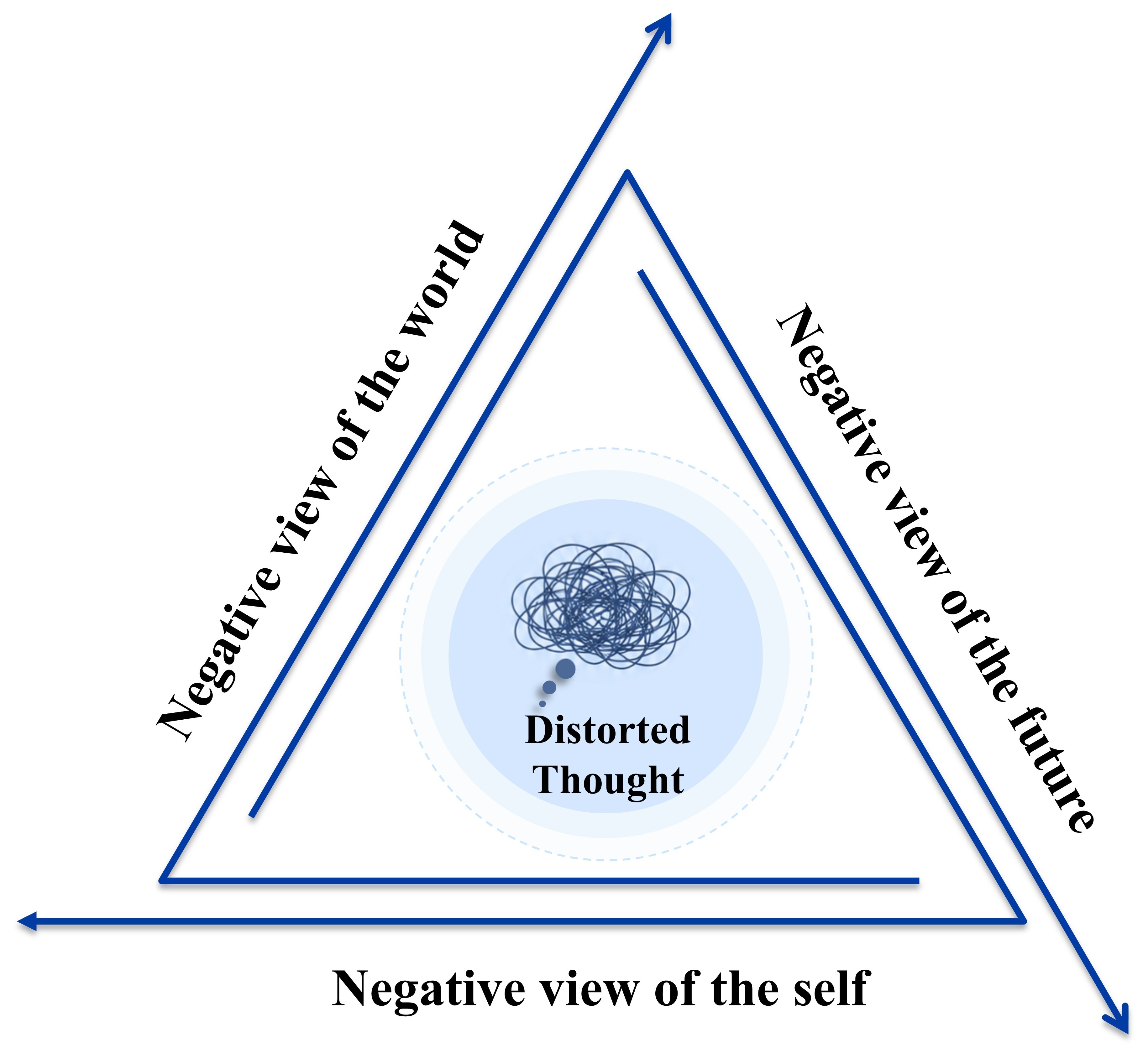}
\caption{Illustration of Beck's cognitive triad, comprising negative views of the self, the world, and the future.}
\label{fig:cognitive-triad}
\end{figure}

Mental disorders have become a global public health challenge, affecting more than one billion individuals worldwide, or approximately one in eight people \citep{who2022,who2025}. Cognitive distortions---systematic biases that negatively skew the interpretation of situations---play a central role in the development and persistence of psychological distress \citep{beck1979,dozois2008}. In cognitive behavioral therapy (CBT), identifying such distortions is a fundamental step toward restructuring maladaptive thought patterns, making automatic cognitive distortion detection important for scalable mental health support \citep{beck1964,beck2020,hofmann2012}.

Recent advances in large language models (LLMs) have enabled increasingly sophisticated natural language reasoning \citep{wei2022,kojima2022,yang2023}, leading to growing interest in cognitive distortion detection \citep{chen2023,nazarova2023,lim2024,lee2024,singh2024,kim2026}. However, many existing approaches still represent distorted thoughts primarily as unstructured text and rely heavily on surface-level linguistic patterns \citep{sage2025}. Such representations provide limited access to the psychological structure underlying distorted thoughts and may overlook how multiple cognitive perspectives jointly characterize an utterance.

This limitation is important because cognitive distortion theory is inherently structured. Beck's cognitive triad characterizes distorted thinking through negative views of the self, the world, and the future, which can interact and reinforce one another \citep{beck1979,beck2014}. Figure~\ref{fig:cognitive-triad} illustrates this theoretical framework. Nevertheless, these perspectives are rarely modeled as explicit and interacting representations in cognitive distortion detection systems.

To address this gap, we propose MTI-GNN (Multi-Perspective Triad Interaction Graph Neural Network). A zero-shot LLM decomposes each utterance into self, world, and future perspectives, after which perspective-specific similarity graphs are constructed for the original text and the extracted triad representations. A Multi-Perspective GNN encodes these graphs, the Triad Interaction module models cross-perspective dependencies through sequential cross-attention, and Prototype-Guided Perspective Fusion performs label-conditioned aggregation. We further adopt label-expanded supervision so that all available distortion annotations contribute to training.

We evaluate MTI-GNN on a multilingual benchmark comprising 9,764 samples from four datasets in Korean, English, and Chinese across ten distortion categories. MTI-GNN significantly outperforms all supervised baselines and exceeds eight generative models evaluated under zero-shot and few-shot prompting. Perspective-removal ablations show that the self, world, and future views each contribute significantly, while evaluation by two licensed psychologists provides preliminary evidence that the extracted representations align with their intended cognitive dimensions.

Our contributions are summarized as follows:

\begin{enumerate}
    \item We introduce a cognitively grounded framework that represents distorted thoughts through the complementary self, world, and future perspectives of Beck's cognitive triad.

    \item We propose MTI-GNN, which combines perspective-specific graph learning, sequential cross-perspective interaction, label-conditioned fusion, and label-expanded supervision.

    \item We conduct a multilingual evaluation against supervised and generative baselines, supported by perspective and interaction-order ablations and human expert validation.
\end{enumerate}

\section{Related Work}

\subsection{Cognitive Distortion Detection}

Early work on cognitive distortion detection relied on rule-based systems using handcrafted linguistic patterns \citep{wiemerhastings2004,bathina2021}. To improve scalability and generalization, later studies combined engineered representations, including TF-IDF and LIWC features, with classifiers such as logistic regression and SVMs \citep{simms2017,shickel2020,shreevastava2021}. However, these approaches remained dependent on manual feature design.

The introduction of distributed word representations enabled neural models to capture richer semantic patterns. Early approaches combined Word2Vec and GloVe embeddings with CNNs and LSTMs \citep{xing2017,rojasbarahona2018,mostafa2021}, while transformer-based models such as BERT further improved performance through contextualized representations and task-specific fine-tuning \citep{devlin2019,lybarger2022,ding2022,ji2022}.

More recently, large language models (LLMs) have enabled cognitive distortion detection through zero-shot and chain-of-thought prompting \citep{chen2023,lim2024}. Subsequent work has addressed cross-dataset generalization through multi-task learning \citep{qi2025} and explored psychologically grounded decomposition of distorted thoughts through multi-view modeling \citep{kim2026}.

Despite these advances, most existing approaches encode each utterance primarily as a single textual sequence, with limited exploration of psychologically structured and interacting representations. This gap motivates the development of representational frameworks that explicitly incorporate the cognitive structure underlying distorted thoughts.

\subsection{Graph-Based Text Classification}

Graph neural networks (GNNs) have been widely applied to text classification by modeling relational structures among documents, words, and labels \citep{kipf2017,velickovic2018}. Early studies such as TextGCN and BertGCN showed that corpus-level graph structures can improve classification by jointly capturing semantic and structural information through graph propagation \citep{yao2019,lin2021}.

Subsequent research extended graph-based frameworks to multi-label classification by incorporating label semantics and dependencies through label-conditioned graph learning, label co-occurrence structures, token--label interactions, and correlation-aware graph decomposition \citep{ma2021,li2022,vu2023,bei2025}.

More recently, integrating large language models with graph neural networks has emerged as a promising direction, combining contextual semantic representations with the structural inductive bias of graph learning for complex reasoning and classification tasks \citep{chengraph2024}.

Motivated by these advances, our approach constructs perspective-specific graphs from the original text and cognitive triad representations. Unlike conventional text graphs organized primarily around documents, words, or labels, MTI-GNN separately encodes similarity graphs constructed from the original text and from each of the self, world, and future perspectives. It then models dependencies among these perspectives through sequential cross-attention and performs label-conditioned fusion, thereby incorporating psychologically grounded structure beyond flat textual representations.

\section{Datasets}
\label{sec:datasets}

\subsection{Dataset Overview}
\label{sec:dataset-overview}

We evaluate our framework on four publicly available cognitive distortion datasets spanning Korean, English, and Chinese: TherapistQA \citep{shreevastava2021}, SocialCD-3K \citep{qisocial2025}, KoACD \citep{koacd2025}, and the Cognitive Reframing Dataset \citep{sharma2023}. The datasets cover patient--therapist interactions, social media posts, synthetic counseling utterances, and situation--negative thought pairs, and include both single-label and multi-label annotation settings. Table~\ref{tab:dataset-overview} summarizes their languages, annotation settings, and sample statistics. Detailed descriptions of the individual datasets are provided in Appendix~\ref{app:dataset-details}.

\subsection{Label Standardization}
\label{sec:label-standardization}

Although the four datasets are broadly grounded in CBT, they adopt partially different cognitive distortion taxonomies. To enable unified training and evaluation, we standardize their annotations into ten categories based on Beck's cognitive therapy framework \citep{beck1979} and Burns's cognitive distortion typology \citep{burns1981}. Semantically overlapping labels are merged according to their theoretical definitions. For example, Mind Reading and Fortune Telling are unified as Jumping to Conclusions, while Catastrophizing is mapped to Magnification. Labels that cannot be reliably aligned with the unified taxonomy, have insufficient samples, or provide limited cross-dataset comparability are excluded. The complete taxonomy, category definitions, mapping rules, and original label distributions are provided in Appendix~\ref{app:dataset-details}.

\begin{table}[t]
\centering
\small
\renewcommand{\arraystretch}{1.2}
\begin{tabular}{ccccc}
\toprule
\textbf{Dataset} & \textbf{Lang} & \textbf{Setting} & \textbf{Samples} & \textbf{Labels} \\
\midrule
TherapistQA & English & Mixed & 1,597 & 1,999 \\
SocialCD-3K & Chinese & Mixed & 3,387 & 3,962 \\
KoACD & Korean & Single & 4,510 & 4,510 \\
\makecell{Cognitive \\ Reframing} & English & Mixed & 270 & 491 \\
\bottomrule
\end{tabular}
\caption{Statistics of the four datasets used in this study after preprocessing and label standardization.}
\label{tab:dataset-overview}
\end{table}

\section{Method}
\label{sec:method}

\subsection{Cognitive Triad Extraction}
\label{sec:triad-extraction}

In cognitive behavioral therapy (CBT), distorted thinking is often characterized by negative beliefs about the self, the world, and the future, collectively known as the cognitive triad \citep{beck1979,beck2014}. Motivated by this framework, we decompose each utterance into three perspectives: self, world, and future. The self perspective reflects beliefs about oneself, the world perspective captures interpretations of external situations or others, and the future perspective represents expectations about future outcomes. Together with the original utterance, they form four complementary views: text, self, world, and future.

We extract the triad perspectives using GPT-4o-mini \citep{openai2024}. Given an utterance, the model generates structured self, world, and future representations and returns an empty value when a perspective is not supported by the input, reducing unsupported inference. Generation settings are provided in Appendix~\ref{app:llm-generation-settings}, and the extraction prompt is reported in Appendix~\ref{app:prompts}. Because these representations form the basis of the perspective-specific graphs, we evaluate their quality through embedding-based analysis, an independent LLM-based assessment, and human expert validation in Section~\ref{sec:model-analysis}.

\begin{figure*}[t]
\centering
\includegraphics[width=\textwidth]{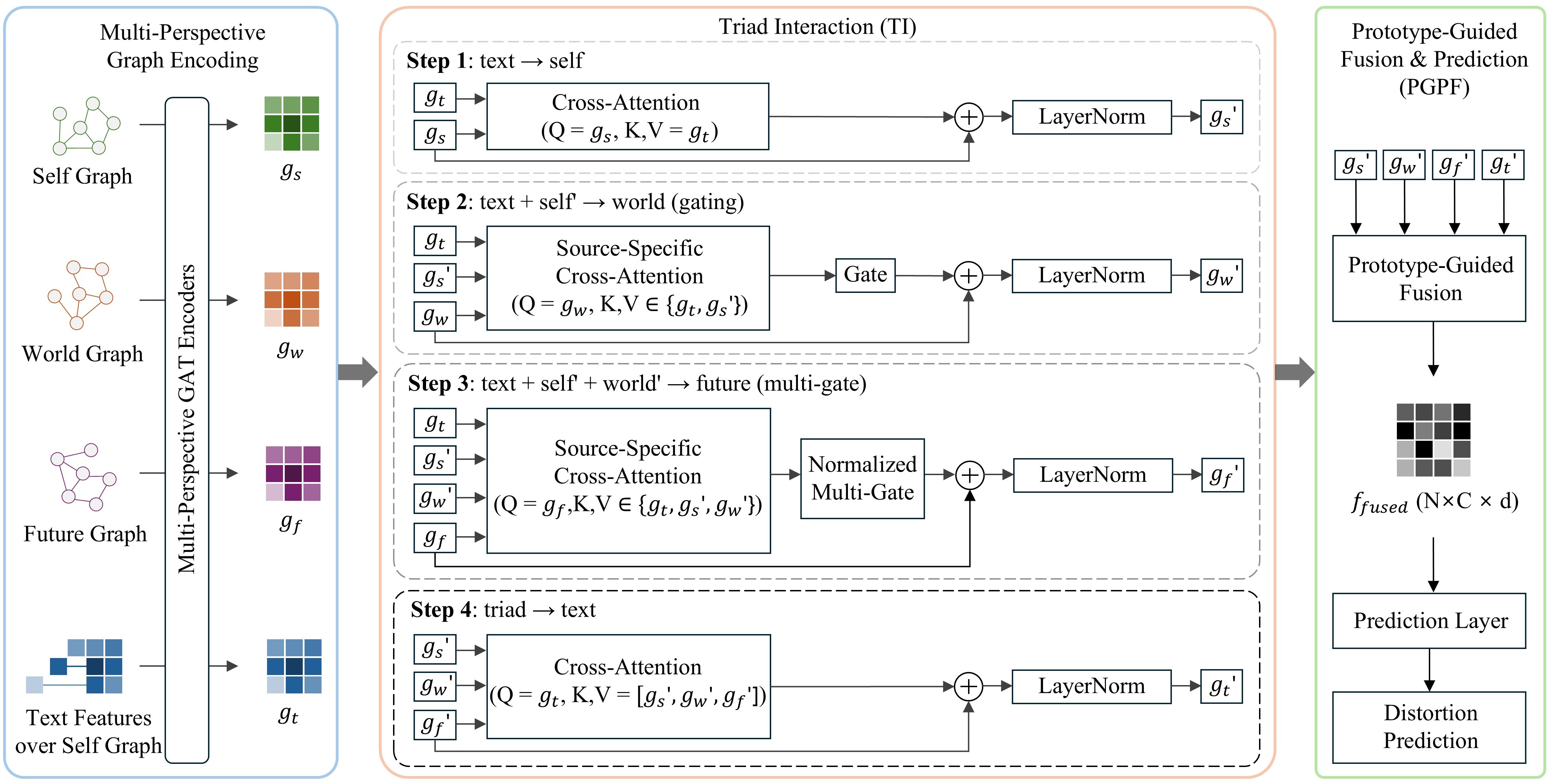}
\caption{Overall architecture of MTI-GNN, consisting of perspective-specific graph encoding, sequential Triad Interaction, and Prototype-Guided Perspective Fusion.}
\label{fig:mti-gnn}
\end{figure*}

\subsection{Multi-Perspective Embedding and Graph Construction}
\label{sec:graph-construction}

We encode the extracted self, world, and future perspectives using BGE-M3, which provides multilingual representations for Korean, English, and Chinese text \citep{chenbge2024}. For each perspective, we compute pairwise cosine similarities between the corresponding utterance representations and construct a directed \(K\)-nearest-neighbor graph, where each utterance is represented as a node with outgoing edges to its \(K\) most similar neighbors. The value of \(K\) is selected through homophily analysis and validation-based ablation, as detailed in Appendix~\ref{app:neighborhood-selection}.

For the text view, we reuse the self-view graph topology while replacing its node features with representations from the fine-tuned text encoder described in Section~\ref{sec:text-encoding}. This design propagates contextual text information through neighborhoods defined by self-related cognitive similarity, while keeping the text and self representations distinct.

We also define fixed semantic prototypes for the ten standardized distortion categories using BGE-M3 embeddings of their canonical definitions. These prototypes do not participate in graph message passing and instead serve as semantic queries in the Prototype-Guided Perspective Fusion layer.

\subsection{Triad-Augmented Text Encoding}
\label{sec:text-encoding}

To incorporate cognitive triad information into the contextual text representation, we concatenate the original utterance with its extracted self, world, and future perspectives using a structured template with special delimiter tokens. Perspectives returned as empty are omitted from the sequence.

We fine-tune XLM-RoBERTa-base \citep{conneau2020} on the augmented inputs using the label-expanded supervision described below. For an utterance with multiple distortion labels, the same input contributes one supervision pair per annotated label, allowing all available annotations to train the encoder without creating additional graph nodes.

The final hidden state of the sequence-initial token is used as the text-view embedding. Unlike the independently encoded triad perspectives, this representation jointly captures the original linguistic context and the extracted cognitive perspectives.

\subsection{MTI-GNN}
\label{sec:mti-gnn}

MTI-GNN consists of three components: perspective-specific graph encoders, a Triad Interaction (TI) module, and Prototype-Guided Perspective Fusion (PGPF). Figure~\ref{fig:mti-gnn} illustrates the overall architecture. For utterance \(i\), we denote the text, self, world, and future representations by \(g_i^t\), \(g_i^s\), \(g_i^w\), and \(g_i^f\), respectively.

\paragraph{Multi-Perspective Graph Encoding.}
For each view \(v\in\{t,s,w,f\}\), we apply an independent Graph Attention Network (GAT) \citep{velickovic2018}. The self, world, and future encoders use their corresponding similarity graphs, whereas the text encoder reuses the self-view graph topology with text-view node features, as described in Section~\ref{sec:graph-construction}. Each encoder consists of a linear projection followed by residual GAT blocks with GELU activation and layer normalization, producing the node representation \(g_i^v\). Separate encoders allow the model to capture relational patterns specific to each cognitive perspective.

\paragraph{Triad Interaction.}
The graph encoders initially process the four views independently, although Beck's cognitive triad suggests that the self, world, and future perspectives can interact and reinforce one another \citep{beck1979}. TI therefore updates the triad perspectives sequentially in the self--world--future order using learned source projections and feature-wise gating, followed by triad-to-text cross-attention.

The self view first receives a learned text-derived update, followed by a residual connection and layer normalization, without an additional gating parameter. The world view then receives separate learned projections of the text view and the updated self view, producing source-specific updates \(\Delta_{i,w}^{t}\) and \(\Delta_{i,w}^{s}\). These outputs are combined by the feature-wise complementary gate
\(\gamma_{i,w}=\sigma(W_w[\Delta_{i,w}^{t};\Delta_{i,w}^{s}]+b_w)\), yielding
\(\gamma_{i,w}\odot\Delta_{i,w}^{t}+(1-\gamma_{i,w})\odot\Delta_{i,w}^{s}\).

Finally, the future view receives projected representations from the text and the updated self and world views. For each source \(j\in\{t,s,w\}\), TI computes a projected source representation \(\Delta_{i,f}^{j}=\phi_{f,j}(h_i^j)\) and a target-conditioned feature-wise gate
\(a_{i,f}^{j}=\sigma(W_{f,j}[g_i^f;\Delta_{i,f}^{j}]+b_{f,j})\).
The gates are normalized across sources and used to update the future view:
\begin{equation}
\begin{aligned}
\lambda_{i,f}^{j}
&=
\frac{a_{i,f}^{j}}
{\sum_{r\in\{t,s,w\}}a_{i,f}^{r}+\epsilon},\\
g_i^{f\prime}
&=
\mathrm{LN}\left(
g_i^f+
\sum_{j\in\{t,s,w\}}
\lambda_{i,f}^{j}\odot\Delta_{i,f}^{j}
\right).
\end{aligned}
\label{eq:future-interaction}
\end{equation}
Here, \(h_i^t=g_i^t\), \(h_i^s=g_i^{s\prime}\), and \(h_i^w=g_i^{w\prime}\), while \(\phi_{f,j}\) denotes the learned value and output projections for source \(j\).

After the triad perspectives are updated, they are concatenated along the sequence dimension, and the original text representation attends over the resulting three-element triad sequence. The resulting triad-conditioned update is added to the text view through a residual connection and layer normalization. Cognitive triad theory does not prescribe a fixed causal or computational order \citep{beck1979,beck2014}; therefore, self--world--future is treated as an architectural inductive bias rather than a causal claim. Each triad perspective is updated once, and later perspectives do not revise earlier ones. The final triad-to-text attention serves as terminal aggregation rather than reverse revision among the triad perspectives. We evaluate all six forward orderings in Section~\ref{sec:interaction-order}.

\paragraph{Prototype-Guided Perspective Fusion.}
For each standardized distortion category \(c\), we define a fixed prototype \(p_c\) by encoding its canonical definition with BGE-M3. Each prototype serves as a semantic query over the updated views
\(\tilde{g}_i^v\in\{g_i^{t\prime},g_i^{s\prime},g_i^{w\prime},g_i^{f\prime}\}\).
The query is computed as \(q_c=W_q(W_p p_c)\), while each attention key is obtained by applying \(W_k\) to the \(\ell_2\)-normalized view representation. The original, unnormalized views are retained as values without a separate value projection:
\begin{equation}
\begin{aligned}
\alpha_{i,c}^v
&=
\mathrm{softmax}_v
\left(
\frac{
q_c^\top
W_k\!\left(
\tilde{g}_i^v/\|\tilde{g}_i^v\|_2
\right)}
{\sqrt{d}}
\right),\\
z_{i,c}
&=
\sum_v
\alpha_{i,c}^v\tilde{g}_i^v.
\end{aligned}
\label{eq:prototype-fusion}
\end{equation}
A separate two-layer MLP with GELU activation and dropout maps each label-specific representation \(z_{i,c}\) to its corresponding logit.

\paragraph{Training Objective.}
To use all available distortion annotations, we construct a label-expanded supervision set. Let \(\mathcal{D}_{\mathrm{train}}\) be the original training set and \(Y_i\) the set of labels assigned to utterance \(i\):
\begin{equation}
\mathcal{D}_{\mathrm{train}}^{+}
=
\left\{
(i,c)
\mid
i\in\mathcal{D}_{\mathrm{train}},
\ c\in Y_i
\right\}.
\label{eq:expanded-training}
\end{equation}

A single-label utterance contributes one supervision pair, whereas a multi-label utterance contributes one pair per annotated label. The same utterance and graph node representations are reused across these pairs, so label expansion does not create additional graph nodes. We optimize the model using class-weighted cross-entropy over the expanded pairs:
\begin{equation}
\mathcal{L}
=
-
\frac{1}{
|\mathcal{D}_{\mathrm{train}}^{+}|
}
\sum_{(i,c)\in\mathcal{D}_{\mathrm{train}}^{+}}
w_c
\log P_{\theta}(c\mid i),
\label{eq:training-loss}
\end{equation}
where \(w_c\) denotes the class weight for distortion category \(c\).

\section{Experiments}
\label{sec:experiments}

\subsection{Experimental Setup}
\label{sec:experimental-setup}

\paragraph{Data Split and Training Labels.}
After label standardization, each dataset is stratified by its primary distortion label and split into training, validation, and test sets at an 8:1:1 ratio, after which the corresponding splits are aggregated. During training, each multi-label utterance contributes one supervision pair per annotation while remaining a single graph node; validation and test instances use their primary labels as targets. Table~\ref{tab:data-split} summarizes the resulting split.

\begin{table}[t]
\renewcommand{\arraystretch}{1.2}
\centering
\small
\begin{tabularx}{\linewidth}{l >{\raggedleft\arraybackslash}X}
\toprule
\textbf{Split} & \textbf{Count (\%)} \\
\midrule
Train & 7,804 (79.9\%) \\
Validation & 978 (10.0\%) \\
Test & 982 (10.1\%) \\
\midrule
\textbf{Total} & \textbf{9,764 (100.0\%)} \\
\bottomrule
\end{tabularx}
\caption{Dataset split statistics after label standardization.}
\label{tab:data-split}
\end{table}

\paragraph{Supervised Variants and Component Analysis.}
We compare MTI-GNN with three incremental supervised variants: Text-only MLP uses the original utterance, Text+Triad MLP adds the extracted triad perspectives, and MP-GNN further introduces multi-perspective graph encoding without TI. These variants isolate the contributions of triad augmentation, graph learning, and cross-perspective interaction under identical data splits and label-expanded supervision.

\paragraph{Generative Model Evaluation.}
To provide a strong reference point against recent generative models for unified cognitive distortion classification, we evaluate GPT-4o-mini~\citep{openai2024}, GPT-5.4-mini~\citep{openai2026gpt54mini}, Claude Sonnet 5~\citep{anthropic2026sonnet5}, Qwen3.5-9B~\citep{qwen3.5}, Qwen3.6-27B~\citep{qwen3.6-27b}, Gemma-4-31B~\citep{gemma2026gemma4}, and two GLM variants, GLM-4-9B and GLM-4-32B~\citep{glm2024chatglm}. Each model is tested under text-only and text+triad inputs in both zero-shot and few-shot settings. All models predict one category from the unified ten-label taxonomy. Prompts, demonstrations, decoding settings, and output normalization are provided in Appendix~\ref{app:prompts}.

\paragraph{Implementation and Evaluation.}
Supervised models use XLM-RoBERTa-base as the text encoder and BGE-M3 for triad and prototype embeddings. They are optimized with AdamW and early stopping on validation Weighted-F1. We report test-set Weighted-F1, averaging supervised results over 10 matched random seeds with standard deviations, and assess significance using paired two-tailed \(t\)-tests. Hyperparameters and graph neighborhood selection are detailed in Appendices~\ref{app:implementation-details} and~\ref{app:neighborhood-selection}.

\paragraph{Triad Extraction Validation.}
We validate the extracted perspectives using Gemini 2.5 Flash~\citep{comanici2025} and two licensed clinical psychologists. Gemini scores the self, world, and future representations, including \texttt{none} outputs, on a five-point scale. The experts apply the same criteria to a stratified sample of 150 utterances, resolving disagreements through discussion. This validation is independent of model training and evaluation. Results and protocols are provided in Section~\ref{sec:triad-validation} and Appendices~\ref{app:prompts} and~\ref{app:expert-evaluation}.

\paragraph{Architectural Analyses.}
To examine the architectural role of interaction order and the contribution of each cognitive perspective, we conduct two additional analyses. First, we evaluate all six permutations of the self, world, and future interaction order while keeping the triad-to-text refinement unchanged. Second, we remove each triad perspective individually from the triad-augmented text input, graph encoding, TI, and PGPF. All variants use matched splits, training settings, and random seeds, and are compared with the full model using paired two-tailed \(t\)-tests. Results are presented in Section~\ref{sec:main-results}.

\begin{table}[t]
\renewcommand{\arraystretch}{1.15}
\centering
\small
\begin{tabular*}{\linewidth}{@{\extracolsep{\fill}}lc@{}}
\toprule
\textbf{Model / Dataset} & \textbf{Weighted-F1} \\
\midrule
\multicolumn{2}{l}{\textit{Supervised model comparison}} \\
Text-only MLP & \(0.5161 \pm 0.0042\) \\
Text+Triad MLP & \(0.5387 \pm 0.0047\) \\
MP-GNN & \(0.5429 \pm 0.0024\) \\
\textbf{MTI-GNN} & \(\mathbf{0.5555 \pm 0.0015}\) \\
\midrule
\multicolumn{2}{l}{\textit{MTI-GNN by dataset}} \\
\quad TherapistQA & \(0.4726 \pm 0.0118\) \\
\quad KoACD & \(0.5344 \pm 0.0054\) \\
\quad Cognitive Reframing & \(0.4579 \pm 0.0211\) \\
\quad SocialCD-3K & \(0.6574 \pm 0.0091\) \\
\bottomrule
\end{tabular*}
\caption{Supervised model comparison and dataset-level performance of MTI-GNN under label-expanded training. Values are mean \(\pm\) standard deviation over 10 runs. MTI-GNN significantly outperforms each supervised variant (\(p<0.001\), paired two-tailed \(t\)-test).}
\label{tab:supervised-results}
\end{table}

\begin{table*}[t]
\centering
\small
\renewcommand{\arraystretch}{1.15}
\begin{tabular*}{\textwidth}{@{\extracolsep{\fill}}lcccc@{}}
\toprule
\textbf{Model}
& \textbf{Text Zero-Shot}
& \textbf{Text Few-Shot}
& \textbf{Text+Triad Zero-Shot}
& \textbf{Text+Triad Few-Shot} \\
\midrule
GPT-4o-mini & 0.3814 & 0.3809 & 0.3875 & 0.3872 \\
GPT-5.4-mini & 0.4202 & 0.4194 & 0.4190 & 0.4232 \\
Claude Sonnet 5 & 0.4078 & 0.4126 & 0.4180 & 0.4288 \\
Qwen3.5-9B & 0.3719 & 0.3822 & 0.3797 & 0.3787 \\
Qwen3.6-27B & 0.4118 & 0.4110 & \(\mathbf{0.4359}\) & 0.4348 \\
Gemma-4-31B & 0.4223 & 0.4256 & 0.4206 & 0.4341 \\
GLM-4-9B & 0.3770 & 0.3665 & 0.3532 & 0.3841 \\
GLM-4-32B & 0.4009 & 0.4238 & 0.3981 & 0.3973 \\
\bottomrule
\end{tabular*}
\caption{Weighted-F1 of eight generative models under text-only and triad-augmented zero-shot and few-shot prompting. Bold indicates the best generative-model result.}
\label{tab:generative-results}
\end{table*}

\subsection{Main Results and Ablation Studies}
\label{sec:main-results}

\paragraph{Supervised Performance.}

As shown in Table~\ref{tab:supervised-results}, MTI-GNN achieves the strongest overall performance and significantly outperforms all supervised variants. The stepwise improvements support the contributions of triad augmentation, perspective-specific graph encoding, and cross-perspective interaction. Dataset-level results vary substantially, consistent with previously reported cross-dataset heterogeneity in cognitive distortion detection \citep{qi2025}. Differences in language, source domain, annotation practices, and test-set size may contribute to this variation.

\paragraph{Comparison with Generative Models.}
\label{sec:generative-comparison}

Table~\ref{tab:generative-results} shows that MTI-GNN exceeds all eight generative models across the evaluated prompting settings. Larger variants consistently outperform their smaller counterparts within the evaluated Qwen and GLM model families, whereas few-shot demonstrations and triad-augmented prompting do not yield consistent improvements across models. Nevertheless, all prompted generative models remain below the task-specific supervised MTI-GNN on this benchmark. This comparison is limited to prompted zero-shot and few-shot settings.

\paragraph{Perspective Contributions.}
\label{sec:perspective-ablation}

\begin{table}[t]
\centering
\small
\renewcommand{\arraystretch}{1.15}
\begin{tabular*}{\linewidth}{@{\extracolsep{\fill}}lcc@{}}
\toprule
\textbf{Removed View}
& \textbf{Weighted-F1}
& \textbf{\(p\)-value} \\
\midrule
None (Full Model) & \(\mathbf{0.5555 \pm 0.0015}\) & -- \\
Self & \(0.5460 \pm 0.0042\) & \(0.000081\) \\
World & \(0.5404 \pm 0.0102\) & \(0.001376\) \\
Future & \(0.5402 \pm 0.0080\) & \(0.000341\) \\
\bottomrule
\end{tabular*}
\caption{Leave-one-perspective-out ablation of MTI-GNN. The \(p\)-values compare each variant with the full model using paired two-tailed \(t\)-tests over matched random seeds.}
\label{tab:perspective-ablation}
\end{table}

Table~\ref{tab:perspective-ablation} shows that removing any perspective significantly reduces performance, confirming that the self, world, and future views provide complementary information. The larger reductions after removing the world or future views indicate that these perspectives contribute more strongly under the proposed interaction architecture.

\paragraph{Interaction Order.}
\label{sec:interaction-order}

\begin{table}[t]
\centering
\small
\renewcommand{\arraystretch}{1.15}
\begin{tabular*}{\linewidth}{@{\extracolsep{\fill}}lr@{}}
\toprule
\textbf{Interaction Order} & \textbf{Weighted-F1} \\
\midrule
Self \(\rightarrow\) World \(\rightarrow\) Future
& \(\mathbf{0.5555 \pm 0.0015}\) \\
World \(\rightarrow\) Future \(\rightarrow\) Self
& \(0.5501 \pm 0.0060\) \\
Future \(\rightarrow\) World \(\rightarrow\) Self
& \(0.5479 \pm 0.0045\) \\
Future \(\rightarrow\) Self \(\rightarrow\) World
& \(0.5478 \pm 0.0056\) \\
World \(\rightarrow\) Self \(\rightarrow\) Future
& \(0.5476 \pm 0.0077\) \\
Self \(\rightarrow\) Future \(\rightarrow\) World
& \(0.5471 \pm 0.0063\) \\
\bottomrule
\end{tabular*}
\caption{Weighted-F1 across triad interaction orders. Text conditioning and triad-to-text refinement are fixed across variants. Values are mean \(\pm\) standard deviation over matched random seeds.}
\label{tab:interaction-order}
\end{table}

Table~\ref{tab:interaction-order} shows that the adopted self--world--future ordering achieves the strongest performance and significantly outperforms all alternatives (\(p<0.05\)). The differences across permutations indicate that MTI-GNN is sensitive to the direction of sequential interaction, rather than merely to the presence of cross-perspective connections. This supports the selected sequence as an empirical architectural inductive bias rather than a causal processing order.

\subsection{Representation and Model Analysis}
\label{sec:model-analysis}

We assess the extracted cognitive representations through embedding similarity, LLM evaluation, and human expert validation, and analyze how MTI-GNN uses the resulting perspectives.

\paragraph{Cognitive Triad Representation Validation.}
\label{sec:triad-validation}

We evaluate the extracted cognitive triad representations using three complementary measures: semantic similarity to assess grounding in the original utterances, and evaluations by both an independent LLM and two licensed clinical psychologists to assess alignment with the self, world, and future dimensions.

\begin{table}[t]
\centering
\small
\renewcommand{\arraystretch}{1.12}
\setlength{\tabcolsep}{2.5pt}
\begin{tabular*}{\linewidth}{@{\extracolsep{\fill}}lcccc@{}}
\toprule
\textbf{Evaluation}
& \textbf{Self}
& \textbf{World}
& \textbf{Future}
& \textbf{Comb.} \\
\midrule
Similarity mean
& 0.669
& 0.693
& 0.661
& 0.764 \\
Similarity std.
& 0.093
& 0.111
& 0.097
& 0.095 \\
None rate (\%)
& 13.4
& 27.3
& 36.4
& 0.8 \\
LLM score
& 4.256
& 4.051
& 3.874
& -- \\
Human score\\[-1mm]
\scriptsize (Consensus)
& 3.93
& 4.19
& 4.04
& 4.05 \\
\bottomrule
\end{tabular*}
\caption{Validation of extracted cognitive triad representations. Similarity is measured against the original utterance, and LLM and consensus human scores use five-point scales.}
\label{tab:triad-validation}
\end{table}

Table~\ref{tab:triad-validation} presents the validation results. Similarity is measured against the original utterance, while the LLM and human consensus scores use five-point scales. Overall, the extracted perspectives remain semantically grounded in the original utterances and jointly capture complementary information. The higher \texttt{none} rates for the world and future perspectives indicate that these dimensions are less explicitly expressed in many utterances, reflecting the sparsity of certain cognitive triad components rather than extraction failure.

To evaluate expert alignment, two licensed clinical psychologists rated 150 representative utterances stratified across datasets and distortion categories. Initial quadratic weighted Cohen's \(\kappa\) values were .339, .276, and .322 for the self, world, and future dimensions, respectively, and were attenuated by differences in rater severity \citep{feinstein1990high}. Joint review of major discrepancies, defined as score differences of at least two points, yielded final consensus scores with an overall mean of 4.05 out of 5.0 (Table~\ref{tab:triad-validation}), supporting alignment with the intended cognitive dimensions.

\paragraph{View Attention Analysis.}

\begin{table}[t]
\renewcommand{\arraystretch}{1.2}
\centering
\small
\begin{tabular*}{\linewidth}{@{\extracolsep{\fill}}lcc@{}}
\toprule
\textbf{View} & \textbf{MP-GNN} & \textbf{MTI-GNN} \\
\midrule
Text & \(0.386 \pm 0.078\) & \(0.257 \pm 0.005\) \\
Self & \(0.222 \pm 0.013\) & \(0.252 \pm 0.003\) \\
World & \(0.200 \pm 0.030\) & \(0.249 \pm 0.003\) \\
Future & \(0.192 \pm 0.038\) & \(0.243 \pm 0.004\) \\
\bottomrule
\end{tabular*}
\caption{View-attention weights of MP-GNN and MTI-GNN, as mean \(\pm\) standard deviation over 10 runs.}
\label{tab:view-attention}
\end{table}

Table~\ref{tab:view-attention} shows that MTI-GNN produces a more balanced attention distribution across the text and cognitive triad perspectives than MP-GNN. MP-GNN places the greatest weight on the text view, whereas MTI-GNN assigns similar weights to all four views, suggesting that TI promotes more even use of the structured cognitive perspectives. However, attention magnitude should not be interpreted as a direct measure of functional contribution, as the leave-one-out results show larger performance reductions for the world and future views despite their similar attention weights.

\paragraph{Label-Wise Performance Analysis.}

\begin{table}[t]
\centering
\small
\renewcommand{\arraystretch}{1.15}
\begin{tabular*}{\linewidth}{@{\extracolsep{\fill}}lc@{}}
\toprule
\textbf{Label} & \textbf{F1} \\
\midrule
All-or-Nothing Thinking
& \(0.4279 \pm 0.0111\) \\
Discounting the Positive
& \(0.7808 \pm 0.0122\) \\
Emotional Reasoning
& \(0.3147 \pm 0.0132\) \\
Jumping to Conclusions
& \(0.5859 \pm 0.0066\) \\
Labeling
& \(0.7430 \pm 0.0082\) \\
Magnification
& \(0.3603 \pm 0.0183\) \\
Mental Filter
& \(0.3720 \pm 0.0142\) \\
Overgeneralization
& \(0.3764 \pm 0.0090\) \\
Personalization
& \(0.5830 \pm 0.0132\) \\
Should Statements
& \(0.7745 \pm 0.0147\) \\
\bottomrule
\end{tabular*}
\caption{Label-wise F1 of MTI-GNN, reported as mean \(\pm\) standard deviation over 10 runs.}
\label{tab:label-results}
\end{table}

Table~\ref{tab:label-results} shows that MTI-GNN achieves the highest label-wise F1 on Discounting the Positive, Should Statements, and Labeling, whose linguistic realizations are comparatively explicit. Emotional Reasoning remains the most difficult category, while Magnification, Mental Filter, and Overgeneralization also show lower performance, reflecting their implicit or semantically overlapping expressions.

\section{Conclusion and Future Work}
\label{sec:conclusion}

We proposed MTI-GNN, a cognitive distortion detection framework grounded in Beck's cognitive triad. It represents utterances through text, self, world, and future views, encodes perspective-specific graphs, models cross-perspective dependencies through Triad Interaction, and performs label-conditioned fusion.

Across four Korean, English, and Chinese datasets, MTI-GNN significantly outperforms all supervised variants and exceeds eight generative models under zero-shot and few-shot prompting. LLM and human expert evaluations provide preliminary evidence that the extracted perspectives align with the intended cognitive dimensions. Ablations confirm the contributions of triad augmentation, graph learning, cross-perspective interaction, and all three views, while interaction-order analysis supports self--world--future as an architectural rather than causal order. Future work will explore dataset-aware adaptation, label co-occurrence modeling, stronger generative baselines, and bidirectional perspective interaction.

\section*{Limitations}
\label{sec:limitations}

Despite the improvements achieved by MTI-GNN, several limitations remain.

\paragraph{Dataset Heterogeneity.}
Performance varies across the four source datasets despite label standardization. Differences in language, domain, annotation practices, and linguistic realization therefore remain important sources of variation, consistent with prior findings \citep{qi2025}. The aggregate results should not be interpreted as uniform cross-domain generalization. Dataset-aware training and domain adaptation remain important directions for future work.

\paragraph{Labeling Scope.}
Label-expanded supervision allows all annotated distortions to contribute to training, but treats each label as an independent target and does not explicitly model co-occurrence dependencies. Validation and test sets also retain primary-label evaluation, so the current results do not establish comprehensive multi-label prediction. Moreover, labels with insufficient samples or limited cross-dataset comparability were excluded during standardization, reducing taxonomy coverage. Larger multi-label datasets and label-dependency-aware objectives are needed to address these limitations.

\paragraph{Triad Extraction Quality.}
Triad perspectives are extracted by a zero-shot LLM, and expert validation covers a 150-utterance subsample rather than the full corpus. The extracted representations should therefore be understood as computational approximations of the cognitive triad rather than clinically verified constructs, and larger-scale expert annotation would strengthen this grounding.

\paragraph{Inter-Rater Reliability in Expert Validation.}
One limitation concerns the inter-rater reliability of the expert validation. Although joint consensus review reconciled major discrepancies, the initial quadratic weighted Cohen's \(\kappa\) values ranged from .276 to .339. This modest agreement reflects the interpretive subjectivity involved in assigning cognitive perspectives to natural-language utterances. Future work will involve a larger expert pool and more granular rating guidelines to improve the robustness of cross-rater validation.

\paragraph{Scope of Generative Model Comparison.}
Our generative-model comparison is limited to zero-shot and few-shot prompting and does not include parameter-efficient or full fine-tuning. The results therefore establish an advantage over prompted generative models under the evaluated settings, rather than over all possible training regimes. Future comparisons should use matched supervision, evaluation protocols, and computational budgets.

\paragraph{Interpretability and Interaction Structure.}
MTI-GNN provides partial interpretability through label-conditioned attention and perspective ablations, but does not generate instance-level reasoning traces. Attention weights should not be interpreted as direct measures of functional contribution. In addition, TI uses an asymmetric single-pass structure in which earlier perspectives are not revised using later ones. Future work should explore explicit explanations and bidirectional or iterative interactions without excessive computational overhead.

\section*{Ethical Considerations}
\label{sec:ethics}

All datasets used in this study are publicly available and were obtained from previously published sources. No additional personal data were collected from the individuals represented in the datasets. Although the original datasets were released under their own data-handling and privacy procedures, we additionally reviewed the data to identify potentially sensitive or personally identifying content before use. No directly identifiable personal information was intentionally retained or introduced during preprocessing or experimentation. The datasets were used solely for research purposes in accordance with their respective terms of use.

To validate the extracted representations, two licensed clinical psychologists evaluated an anonymized subsample of 150 utterances. Both experts participated voluntarily after receiving a full explanation of the study purpose and procedures. Each expert received 45,000 KRW for evaluating 150 items, calculated proportionally from a standard professional rate of 300,000 KRW per 1,000 items. No personally identifiable information was collected beyond records required for compensation.

All evaluation materials were anonymized before review. The consensus procedure was used solely to establish final validation scores and did not affect the extracted representations or downstream experiments. The study analyzed only anonymized secondary text data and involved no direct interaction with data subjects, clinical diagnosis, or therapeutic intervention.

The proposed cognitive distortion detection framework is intended solely for research purposes and should not be considered a substitute for clinical diagnosis, psychological assessment, or therapeutic intervention. Model predictions may be inaccurate or biased, particularly because the framework relies in part on LLM-generated cognitive triad representations. Consequently, model outputs should not be used for clinical decision-making without supervision from qualified mental health professionals.

Cognitive distortion detection inherently involves sensitive mental health-related content, and incorrect predictions or inappropriate deployment may negatively affect vulnerable individuals. Although the extracted perspectives were assessed by licensed clinical psychologists on a representative subsample, they remain computational representations designed for classification rather than comprehensive clinical assessments. Any real-world deployment of such systems therefore requires careful ethical review, broader clinical validation, rigorous performance assessment, privacy safeguards, and appropriate professional oversight to ensure safe and responsible use.

To facilitate manuscript preparation, AI-based assistants were used in a limited capacity for language polishing, readability improvement, and wording refinement. AI assistance was also used to support the translation and interpretation of non-English source materials during dataset understanding. All scientific claims, experimental procedures, analyses, and final manuscript content were reviewed and approved by the authors.

\bibliography{custom}
\clearpage

\appendix

\section{Implementation and Hyperparameter Details}
\label{app:implementation-details}

\subsection{Supervised Model Settings}

Supervised experiments were conducted on a machine equipped with an AMD Ryzen 9 7900X processor and an NVIDIA GeForce RTX 4080 SUPER GPU. The implementation uses PyTorch 2.6.0 with CUDA 12.6 and PyTorch Geometric 2.7.0.

Table~\ref{tab:mti-hyperparameters} summarizes the final MTI-GNN configuration selected based on validation Weighted-F1. The graph neighborhood size was set to \(K=10\) following the homophily and validation analyses described in Appendix~\ref{app:neighborhood-selection}. The selected configuration was fixed across all random seeds, supervised baselines, and ablation experiments.

\begin{table}[t]
\centering
\small
\renewcommand{\arraystretch}{1.15}
\begin{tabular*}{\linewidth}{@{\extracolsep{\fill}}lc@{}}
\toprule
\textbf{Hyperparameter} & \textbf{Value} \\
\midrule
Hidden dimension & 128 \\
GAT layers & 1 \\
GAT attention heads & 4 \\
TI attention heads & 8 \\
Graph neighborhood size \(K\) & 10 \\
Dropout & 0.2 \\
Learning rate & \(1\times10^{-3}\) \\
Optimizer & AdamW \\
Weight decay & \(1\times10^{-4}\) \\
Early stopping patience & 30 \\
Maximum epochs & 100 \\
\bottomrule
\end{tabular*}
\caption{Final hyperparameter settings for MTI-GNN.}
\label{tab:mti-hyperparameters}
\end{table}

\subsection{LLM Generation Settings}
\label{app:llm-generation-settings}

Closed-source models, including GPT-4o-mini, GPT-5.4-mini, and Claude Sonnet 5, were accessed through their respective APIs. Open-weight models, including Qwen3.5-9B, Qwen3.6-27B, Gemma-4-31B, GLM-4-9B, and GLM-4-32B, were evaluated locally on NVIDIA H200 GPUs using the Transformers library.

Table~\ref{tab:llm-decoding-settings} summarizes the temperature settings used for cognitive triad extraction, prompting-based generative-model evaluation, and LLM-based extraction assessment. A temperature of 0.7 is used for triad extraction to allow flexible reformulation of cognitively relevant information while instructing the model to return \texttt{none} when a perspective is unsupported by the utterance. Temperature is set to 0.0 for zero-shot and few-shot classification to minimize decoding variance, and to 0.1 for stable extraction-quality scoring.

\begin{table}[t]
\centering
\small
\renewcommand{\arraystretch}{1.15}
\begin{tabular*}{\linewidth}{@{\extracolsep{\fill}}lc@{}}
\toprule
\textbf{Task} & \textbf{Temperature} \\
\midrule
Cognitive triad extraction & 0.7 \\
Zero-/few-shot classification & 0.0 \\
Triad extraction evaluation & 0.1 \\
\bottomrule
\end{tabular*}
\caption{Temperature settings for LLM-based extraction, classification, and evaluation.}
\label{tab:llm-decoding-settings}
\end{table}

The extracted cognitive triad representations were cached and reused across all supervised and prompting-based experiments to ensure consistent inputs across model comparisons. Detailed extraction, classification, and evaluation prompts are provided in Appendix~\ref{app:prompts}.

\section{Neighborhood Size Selection for Graph Construction}
\label{app:neighborhood-selection}

We select the graph neighborhood size \(K\) through homophily analysis followed by validation-based evaluation. For each candidate \(K\in\{5,7,10,15,20,30,50\}\), we construct the perspective-specific graphs on the training set and compute edge homophily, defined as the proportion of directed edges whose endpoints share the same primary distortion label.

As shown in Table~\ref{tab:graph-homophily}, homophily decreases monotonically as \(K\) increases across all three cognitive perspectives. Although smaller values yield more label-consistent neighborhoods, they also produce sparser connectivity and restrict information propagation. We therefore retain \(K\in\{5,7,10\}\) as validation candidates and select \(K=10\), which achieves the highest validation Weighted-F1 among them.

\begin{table}[t]
\centering
\small
\renewcommand{\arraystretch}{1.15}
\begin{tabular*}{\linewidth}{@{\extracolsep{\fill}}ccccc@{}}
\toprule
\textbf{\(K\)}
& \textbf{Self}
& \textbf{World}
& \textbf{Future}
& \textbf{Average} \\
\midrule
5  & 0.3016 & 0.2139 & 0.1906 & 0.2353 \\
7  & 0.2965 & 0.2119 & 0.1852 & 0.2312 \\
10 & 0.2911 & 0.2078 & 0.1824 & 0.2271 \\
15 & 0.2846 & 0.2012 & 0.1789 & 0.2216 \\
20 & 0.2799 & 0.1978 & 0.1760 & 0.2179 \\
30 & 0.2716 & 0.1917 & 0.1732 & 0.2122 \\
50 & 0.2623 & 0.1849 & 0.1680 & 0.2051 \\
\bottomrule
\end{tabular*}
\caption{Edge homophily across candidate neighborhood sizes \(K\).}
\label{tab:graph-homophily}
\end{table}

\section{Cognitive Distortion Co-occurrence Analysis}
\label{app:cooccurrence-analysis}

\begin{figure*}[t]
\centering
\includegraphics[width=\textwidth]{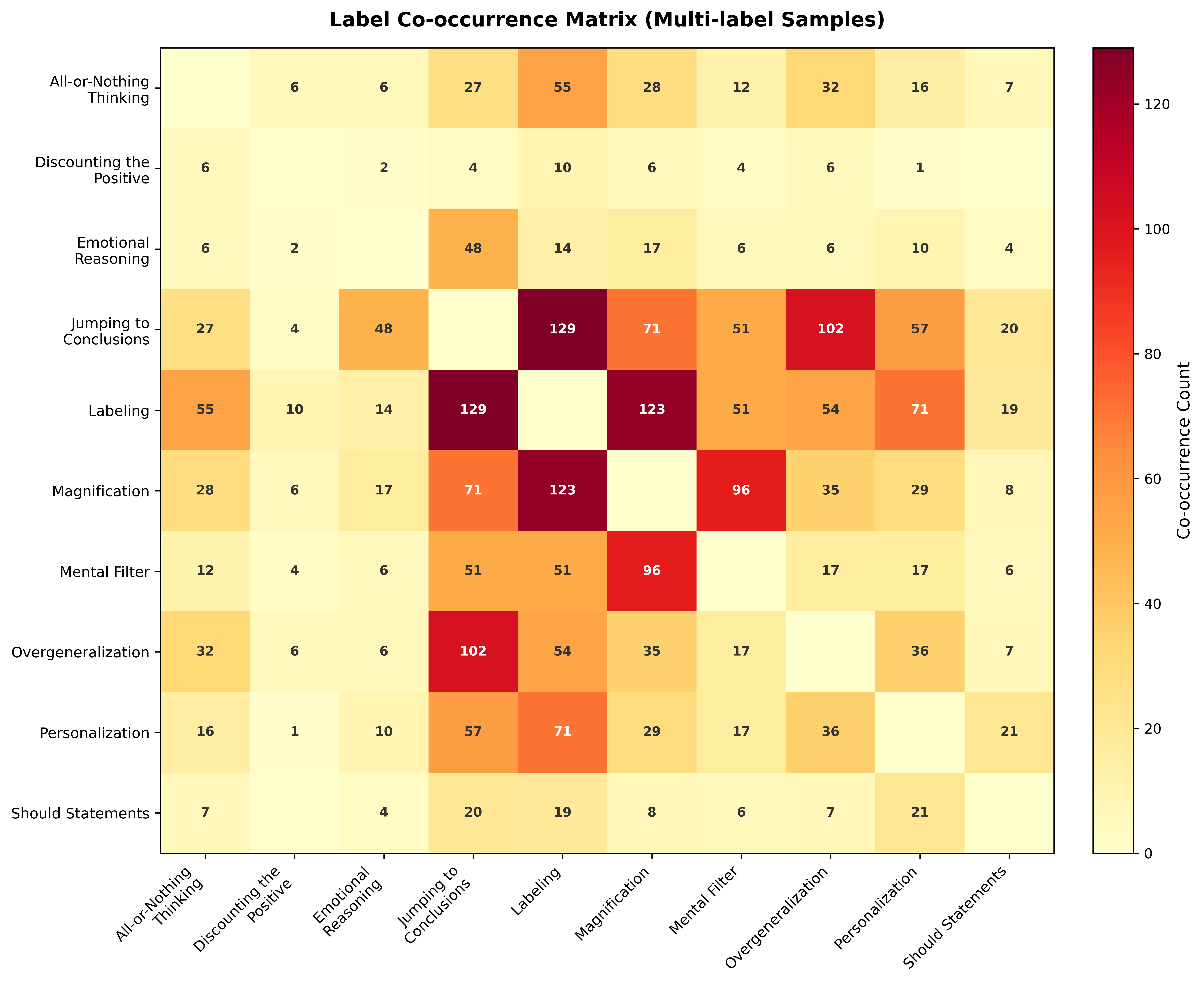}
\caption{Co-occurrence matrix of cognitive distortion labels in the multi-label portion of the aggregated dataset. Darker cells indicate higher co-occurrence frequency.}
\label{fig:cooccurrence-matrix}
\end{figure*}

To examine how distortion categories appear together, we compute a co-occurrence matrix from the 1,073 multi-label samples in the aggregated dataset. For each sample, all annotated labels are considered, and pairwise counts are accumulated across samples.

Figure~\ref{fig:cooccurrence-matrix} shows several notable co-occurrence patterns, particularly among Jumping to Conclusions, Labeling, Magnification, Mental Filter, and Overgeneralization. Frequent pairs include Jumping to Conclusions--Labeling, Labeling--Magnification, and Jumping to Conclusions--Overgeneralization, suggesting that these distortions often appear together within the same utterance.

By contrast, Discounting the Positive and Should Statements show relatively limited co-occurrence with other categories, indicating that they more often appear in isolation. Overall, this analysis highlights the structured nature of multi-label cognitive distortion data and motivates future modeling approaches that explicitly account for co-occurring distortions.

\section{Dataset Details and Label Taxonomy}
\label{app:dataset-details}

\subsection{Source Dataset Descriptions}
\label{app:dataset-descriptions}

\paragraph{TherapistQA.}
TherapistQA contains patient--therapist question--answer interactions distributed through Kaggle. Under the original annotation protocol, each utterance receives one primary distortion label and, when applicable, one additional label \citep{shreevastava2021}.

\paragraph{SocialCD-3K.}
SocialCD-3K contains Chinese social media posts collected from Sina Weibo and manually annotated according to Burns's cognitive distortion framework. Multiple labels may be assigned to a single post \citep{qisocial2025}.

\paragraph{KoACD.}
The Korean Adolescent Cognitive Distortion (KoACD) dataset contains synthetic counseling utterances reconstructed from adolescent counseling posts on NAVER Knowledge iN. Labels were generated and validated through a multi-LLM negotiation framework, with one distortion label assigned to each utterance \citep{koacd2025}.

\paragraph{Cognitive Reframing Dataset.}
The Cognitive Reframing Dataset contains situation--negative thought pairs derived from the Thought Records Dataset and the Mental Health America platform, together with reframing responses written by domain experts. We map its distortion annotations to the unified taxonomy described in Section~\ref{sec:label-standardization}. Most instances contain multiple labels \citep{sharma2023}.

\subsection{Standardized Cognitive Distortion Taxonomy}
\label{app:label-taxonomy}

Table~\ref{tab:distortion-taxonomy} presents the ten cognitive distortion categories used in this study, together with concise definitions and illustrative examples. The taxonomy is grounded in established cognitive behavioral therapy frameworks \citep{beck1979,burns1981}. Semantically overlapping source labels are mapped to common categories according to their theoretical definitions, as described in Section~\ref{sec:label-standardization}.

\begin{table*}[t]
\centering
\footnotesize
\renewcommand{\arraystretch}{1.12}
\setlength{\tabcolsep}{3.5pt}
\begin{tabularx}{\textwidth}{
>{\raggedright\arraybackslash}p{2.5cm}
>{\raggedright\arraybackslash}X
>{\raggedright\arraybackslash}X
}
\toprule
\textbf{Label}
& \textbf{Definition}
& \textbf{Illustrative Example} \\
\midrule

All-or-Nothing Thinking &
Viewing situations in absolute, binary terms without recognizing intermediate possibilities. &
``I missed one question, so I completely failed the exam.'' \\

Overgeneralization &
Drawing a broad conclusion from a single event and expecting the same pattern to recur. &
``I struggled in one presentation, so I will always fail in public.'' \\

Mental Filter &
Focusing selectively on negative details while disregarding positive evidence. &
``Everyone praised my work, but I can think only about one minor criticism.'' \\

Discounting the Positive &
Dismissing positive experiences or achievements as insignificant, accidental, or undeserved. &
``The promotion was just luck; it does not mean I did well.'' \\

Jumping to Conclusions &
Drawing negative conclusions without sufficient evidence, including mind reading and fortune telling. &
``My manager was quiet, so they must be angry with me.'' \\

Magnification &
Exaggerating the importance or consequences of problems, mistakes, or weaknesses. &
``One typo will ruin everyone's trust in me.'' \\

Emotional Reasoning &
Treating an emotional state as evidence that a belief or prediction is true. &
``I feel anxious, so I know I will fail.'' \\

Should Statements &
Imposing rigid expectations about how oneself, others, or the world must behave. &
``I should never feel nervous.'' \\

Labeling &
Defining oneself or others globally based on an isolated behavior or mistake. &
``I missed a deadline, so I am a failure.'' \\

Personalization &
Assuming excessive responsibility for events that are not fully under one's control. &
``My child had a conflict at school, so it must be entirely my fault.'' \\

\bottomrule
\end{tabularx}
\caption{Standardized cognitive distortion categories with concise definitions and author-written illustrative examples.}
\label{tab:distortion-taxonomy}
\end{table*}

\subsection{Original Label Distributions}
\label{app:original-label-distributions}

Table~\ref{tab:original-label-distributions} summarizes the original label distributions of the four source datasets before standardization. Labels that could not be reliably aligned with the unified taxonomy or provided insufficient cross-dataset comparability were excluded from the final experiments, as described in Section~\ref{sec:label-standardization}.

\begin{table*}[t]
\centering
\fontsize{9}{9.5}\selectfont
\renewcommand{\arraystretch}{1.2}
\setlength{\tabcolsep}{5pt}

\begin{tabularx}{\textwidth}{
>{\raggedright\arraybackslash}X
>{\centering\arraybackslash}p{1.7cm}
>{\centering\arraybackslash}p{1.8cm}
>{\centering\arraybackslash}p{1.2cm}
>{\centering\arraybackslash}p{1.6cm}
>{\centering\arraybackslash}p{1.0cm}
}

\toprule

\textbf{Original Label}
&
\textbf{TherapistQA}
&
\textbf{SocialCD-3K}
&
\textbf{KoACD}
&
\textbf{CR Dataset}
&
\textbf{Total}

\\

\midrule

All-or-Nothing Thinking
&
100 & 77 & 464 & 99 & 740
\\

Overgeneralization
&
239 & 141 & 452 & 107 & 939
\\

Mental Filter (Negative Filtering)
&
122 & 378 & 470 & -- & 970
\\

Discounting the Positive (Disqualifying the Positive)
&
-- & 27 & 451 & 40 & 518
\\

Jumping to Conclusions
&
382 & 773 & 431 & 149 & 1,735
\\

Magnification (Magnification and Minimization)
&
195 & 321 & 432 & 83 & 1,031
\\

Emotional Reasoning
&
134 & 16 & 458 & 43 & 651
\\

Should Statements
&
107 & 84 & 415 & 22 & 628
\\

Labeling (Labeling and Mislabeling)
&
165 & 1,961 & 478 & 102 & 2,706
\\

Personalization (Personalizing)
&
153 & 215 & 459 & 98 & 925
\\

No Distortion (None)
&
933 & -- & -- & 19 & 952
\\

Negative Feeling or Emotion
&
-- & -- & -- & 151 & 151
\\

Blaming Oneself
&
-- & 188 & -- & -- & 188
\\

Blaming Others
&
-- & 27 & -- & -- & 27
\\

Blaming
&
-- & -- & -- & 34 & 34
\\

Comparing and Despairing
&
-- & -- & -- & 12 & 12
\\

\midrule

\textbf{Total}
&
\textbf{2,530}
&
\textbf{3,993}
&
\textbf{4,510}
&
\textbf{959}
&
\textbf{11,992}

\\

\bottomrule

\end{tabularx}

\caption{Original Label Distribution Across Datasets}
\label{tab:original-label-distributions}

\end{table*}

\begin{table*}[t]
\centering
\small
\renewcommand{\arraystretch}{1.3}

\begin{tabularx}{\textwidth}{
>{\raggedright\arraybackslash}X
}

\toprule

\textbf{Cognitive Triad Extraction}

\\

\midrule

You are a clinical psychology expert specializing in Cognitive Behavioral Therapy (CBT). \\
Your task is to extract Beck's Cognitive Triad elements from a distorted thought statement. \\
Beck's Cognitive Triad consists of three components:

\\

1. Self: Negative views about oneself. This includes feeling inadequate, incompetent, unworthy, unlovable, or blaming oneself. Also includes doubting one's own abilities or feeling like a burden. Even mild self-criticism or self-doubt counts.

\\

2. World: Negative views about the world, others, or one's environment. This includes perceiving others as judgmental, unsupportive, unfair, hostile, or indifferent. Also includes feeling misunderstood, excluded, or that relationships are difficult. Any negative reference to other people or social situations counts.

\\

3. Future: Negative views about the future or what will happen. This includes expecting bad outcomes, anticipating failure, worrying that things won't improve, fearing consequences, or feeling hopeless. Even short-term negative predictions count (e.g., "I will fail this exam", "They will judge me").

\\
\\

Output format (strict JSON only, no markdown, no explanation):

\\

{"self": "extracted expression or none", "world": "extracted expression or none", "future": "extracted expression or none"}

\\

\bottomrule

\end{tabularx}

\caption{Prompt for Cognitive Triad Extraction.}
\label{tab:triad-extraction-prompt}

\end{table*}

\section{Prompt Templates}
\label{app:prompts}

Table~\ref{tab:triad-extraction-prompt} presents the prompt used for cognitive triad extraction. Given an input utterance, GPT-4o-mini is instructed to extract the self, world, and future perspectives according to Beck's cognitive triad framework and return the result in a structured JSON format. If a particular dimension is not present in the text, the model is instructed to return ``none'' rather than inferring unsupported content.

Table~\ref{tab:triad-evaluation-prompt} presents the prompt used to evaluate the quality of triad extraction. Gemini 2.5 Flash is employed as an independent judge to assess whether each extracted perspective accurately reflects the corresponding cognitive triad dimension. Each dimension is rated on a 1--5 scale, and the appropriateness of None assignments is evaluated separately.

Tables~\ref{tab:zero-shot-prompt} and~\ref{tab:few-shot-prompt} present the zero-shot and few-shot prompts used for the generative-model classification baselines, respectively. Both prompts provide the model with definitions of all ten cognitive distortion categories. The few-shot prompt additionally includes representative examples for each category to guide classification. In both settings, only the original utterance is provided as input, without any cognitive triad information.

\begin{table*}[t]
\centering
\small
\renewcommand{\arraystretch}{1.3}

\begin{tabularx}{\textwidth}{
>{\raggedright\arraybackslash}X
}

\toprule

\textbf{Cognitive Triad Extraction Quality Evaluation}

\\

\midrule

You are an expert in Cognitive Behavioral Therapy (CBT) and Beck's cognitive triad theory. Your task is to evaluate whether the extracted cognitive triad elements correctly represent the cognitive triad dimensions from the original text.

\\

Beck's cognitive triad consists of:

\\

Self: Negative beliefs or perceptions about oneself (e.g., "I am worthless", "I am a failure")

\\

World: Negative interpretations of ongoing experiences and the environment (e.g., "The world is unfair", "Nobody cares about me")

\\

Future: Negative expectations about the future (e.g., "Things will never get better", "I have no hope")

\\

Evaluate the cognitive triad extraction for the following text.

\\

Original text: \{original\}

\\

Extracted cognitive triad:

\\

- Self: \{self\}

\\

- World: \{world\}

\\

- Future: \{future\}

\\

Rate each dimension on a scale of 1--5:

\\

1: Completely wrong or irrelevant

\\

2: Mostly incorrect

\\

3: Partially correct

\\

4: Mostly correct

\\

5: Perfectly correct

\\

Also rate none\_appropriateness (1--5): whether None values are appropriately assigned when the dimension is not present in the text.

\\

1: None is wrongly assigned when content exists

\\

5: None is appropriately assigned only when truly absent

\\

Respond with ONLY this JSON format. Do not include any other text, explanation, or markdown:

\\

\{

\\

\quad "self\_score": <1-5>,

\\

\quad "world\_score": <1-5>,

\\

\quad "future\_score": <1-5>,

\\

\quad "none\_appropriateness": <1-5>,

\\

\quad "overall\_score": <1-5>,

\\

\quad "reasoning": "<one sentence explanation>"

\\

\}

\\

\bottomrule

\end{tabularx}

\caption{Prompt for Cognitive Triad Extraction Quality Evaluation.}
\label{tab:triad-evaluation-prompt}

\end{table*}

\begin{table*}[t]
\centering
\small
\renewcommand{\arraystretch}{1.3}

\begin{tabularx}{\textwidth}{
>{\raggedright\arraybackslash}X
}

\toprule

\textbf{Cognitive Distortion Classification (Zero-shot)}

\\

\midrule
\\
\\

You are a clinical psychologist expert in CBT. Classify the given statement into exactly ONE of these 10 cognitive distortion categories.

\\
\\

1. All-or-Nothing Thinking:

A black-and-white style of thinking where situations are viewed as either completely perfect or total failures, with no middle ground.

\\
\\

2. Discounting the Positive:

Brushing off genuine achievements or good experiences as meaningless or accidental, as though they don't truly count.

\\
\\

3. Emotional Reasoning:

Treating your emotional state as direct evidence of reality, assuming that what you feel must be an accurate reflection of the facts.

\\
\\

4. Jumping to Conclusions:

Forming a negative interpretation of a situation without any concrete evidence, either by assuming what others are thinking or by predicting a bad outcome in advance.

\\
\\

5. Labeling:

Attaching a fixed, sweeping negative identity to yourself or others based on a single action or mistake, rather than addressing the specific behavior.

\\
\\

6. Magnification:

Blowing your flaws and mistakes out of proportion while simultaneously downplaying your strengths and accomplishments.

\\
\\

7. Mental Filter:

Zeroing in on a single negative detail while filtering out all positive aspects, which distorts your overall perception of a situation.

\\
\\

8. Overgeneralization:

Drawing a broad, sweeping conclusion from a single negative event, assuming it will repeat endlessly in the future.

\\
\\

9. Personalization:

Taking excessive personal responsibility for events outside your control, or conversely, deflecting your own role in a problem onto others or external circumstances.

\\
\\

10. Should Statements:

Holding rigid internal rules about how you, others, or the world must behave, and feeling guilt or frustration whenever reality falls short of those standards.

\\
\\

Respond with ONLY the exact category name. No explanation.

\\
\\

Statement: \{cognitive\_distortion\}

\\
\\

Cognitive Distortion:

\\

\bottomrule

\end{tabularx}

\caption{Zero-shot Prompt for Cognitive Distortion Classification}
\label{tab:zero-shot-prompt}

\end{table*}

\begin{table*}[t]
\centering
\small
\renewcommand{\arraystretch}{1.3}

\begin{tabularx}{\textwidth}{
>{\raggedright\arraybackslash}X
}

\toprule

\textbf{Cognitive Distortion Classification (Few-shot)}

\\

\midrule

You are a clinical psychologist expert in CBT. Classify the given statement into exactly ONE of these 10 cognitive distortion categories.

1. All-or-Nothing Thinking: A black-and-white style of thinking where situations are viewed as either completely perfect or total failures, with no middle ground.

2. Discounting the Positive: Brushing off genuine achievements or good experiences as meaningless or accidental, as though they don't truly count.

3. Emotional Reasoning: Treating your emotional state as direct evidence of reality, assuming that what you feel must be an accurate reflection of the facts.

4. Jumping to Conclusions: Forming a negative interpretation of a situation without any concrete evidence, either by assuming what others are thinking or by predicting a bad outcome in advance.

5. Labeling: Attaching a fixed, sweeping negative identity to yourself or others based on a single action or mistake, rather than addressing the specific behavior.

6. Magnification: Blowing your flaws and mistakes out of proportion while simultaneously downplaying your strengths and accomplishments.

7. Mental Filter: Zeroing in on a single negative detail while filtering out all positive aspects, which distorts your overall perception of a situation.

8. Overgeneralization: Drawing a broad, sweeping conclusion from a single negative event, assuming it will repeat endlessly in the future.

9. Personalization: Taking excessive personal responsibility for events outside your control, or conversely, deflecting your own role in a problem onto others or external circumstances.

10. Should Statements: Holding rigid internal rules about how you, others, or the world must behave, and feeling guilt or frustration whenever reality falls short of those standards.

Examples:

[All-or-Nothing Thinking]

- After missing one question on an exam, you think "I completely bombed this — the whole test is ruined"

[Discounting the Positive]

- After earning a promotion, you think "It was just luck — anyone in my position would have gotten the same result"

[Emotional Reasoning]

- "I feel so anxious just thinking about the presentation — that must mean I'm going to fail completely"

[Jumping to Conclusions]

- Your manager stays quiet during a meeting, and you immediately conclude "They must be furious with me" — without any confirmation

[Labeling]

- After forgetting a deadline, you tell yourself "I'm a complete failure"

- After a colleague makes an error, you write them off as "totally incompetent"

[Magnification]

- A single typo in a report leads you to think "This is a catastrophic mistake — everyone will lose trust in me," while months of solid work go unacknowledged in your mind

[Mental Filter]

- Your colleagues broadly praise your work, but one person raises a minor critique — and that one comment is all you can think about for days

[Overgeneralization]

- After stumbling through one presentation, you tell yourself "I always mess up in front of people — I'll never be able to do this right"

[Personalization]

- When your child gets into a conflict at school, you immediately think "This is entirely my fault — I must be a bad parent," without considering other contributing factors

[Should Statements]

- "I should never get nervous."

- "She ought to know better than to treat me that way" — leading to self-blame or resentment when expectations aren't met

Respond with ONLY the exact category name. No explanation.

Statement: \{cognitive\_distortion\}

Cognitive Distortion:

\\

\bottomrule

\end{tabularx}

\caption{Few-shot Prompt for Cognitive Distortion Classification}
\label{tab:few-shot-prompt}

\end{table*}

\clearpage
\onecolumn

\clearpage
\onecolumn

\section{Human Expert Evaluation Form}
\label{app:expert-evaluation}

\begingroup
\centering
\small
\renewcommand{\arraystretch}{1.2}

\begin{tabular}{p{\dimexpr\textwidth-2\tabcolsep\relax}}
\hline
\textbf{Expert Evaluation Form for Cognitive Triad Extraction} \\
\hline

Hello. \\[4pt]

Thank you very much for participating in this study. This task evaluates whether the self, world, and future perspectives automatically extracted from each utterance accurately reflect Beck's cognitive triad. The evaluation file contains the original utterance and the corresponding extracted perspectives. \\[4pt]

\textbf{Cognitive Triad Dimensions} \\[2pt]

\textbf{Self:} Negative beliefs or perceptions about oneself, such as ``I am worthless.'' \\

\textbf{World:} Negative interpretations of other people, experiences, or the surrounding environment, such as ``Nobody cares about me.'' \\

\textbf{Future:} Negative expectations about future outcomes, such as ``Things will never get better.'' \\[4pt]

\textbf{Evaluation Criteria} \\[2pt]

For each utterance, assign a score from 1 to 5 to the extracted self, world, and future perspectives. When a perspective is marked as \texttt{none}, evaluate whether that dimension is genuinely absent from the original utterance. \\[4pt]

5 points: Completely accurate, or the \texttt{none} assignment is fully appropriate. \\

4 points: Mostly accurate, with only minor omissions or imprecision. \\

3 points: Partially accurate, but some important information is missing or ambiguous. \\

2 points: Mostly inaccurate or confused with another cognitive dimension. \\

1 point: Completely inaccurate or irrelevant, or the \texttt{none} assignment clearly omits an expressed dimension. \\[4pt]

\textbf{Additional Guidance} \\[2pt]

Evaluate semantic meaning rather than exact wording. Do not penalize an extraction when the original wording is changed but the underlying negative cognition is accurately preserved. Some utterances may overlap across multiple dimensions; in such cases, do not penalize a reasonable assignment to the dimension carrying the primary meaning. \\[4pt]

Some utterances may contain references to suicidal ideation, self-harm, or other emotionally sensitive experiences. If you experience discomfort or emotional distress, you may take a break, stop the evaluation, or request a schedule adjustment at any time. \\[4pt]

\textbf{Response Fields} \\[2pt]

For each utterance, record only three integer scores: \texttt{self\_score}, \texttt{world\_score}, and \texttt{future\_score}, each ranging from 1 to 5. \\[4pt]

If you have any questions about the criteria or evaluation materials, please contact the research team. \\[4pt]

Thank you. \\
\hline
\end{tabular}

\captionof{table}{Human expert evaluation form for cognitive triad extraction.}
\label{tab:expert-evaluation-form}

\endgroup

\twocolumn

\end{document}